\documentclass[letterpaper, 10 pt, conference]{ieeeconf}  % Comment this line out if you need a4paper
\usepackage{graphicx}
\usepackage{caption}
\usepackage{wrapfig}
\usepackage{amsmath}
\usepackage[hidelinks]{hyperref}
\usepackage{cleveref}
\usepackage[hidelinks]{hyperref}
\usepackage[export]{adjustbox}
\usepackage{float}
\let\labelindent\relax
\usepackage[inline]{enumitem}
\usepackage{booktabs}
\usepackage{amssymb}
\usepackage{multirow}
\usepackage{pifont}
\usepackage{xcolor}
\newcommand{\cmark}{\textcolor{green!60!black}{\ding{51}}}
\newcommand{\xmark}{\textcolor{red}{\ding{55}}}

\IEEEoverridecommandlockouts                              % This command is only needed if 
\newcommand{\ours}{ModPack}

\newcommand{\mypara}[1]{\vspace{0mm}\noindent\textbf{#1}}
\usepackage{dblfloatfix}
\title{\LARGE \bf
\ours:\\ Extensible Teleoperation Interface for Bimanual Mobile Manipulation
}

\author{\vspace{-2mm}
  Joshua Citron \qquad Renee Zbizika \qquad Zeyi Liu \qquad Shuran Song \\\\
  \vspace{-2mm}
  \mdseries{Stanford University} \\\\
  \mdseries{\href{https://modpack-robotics.github.io}{\textcolor[RGB]{93,64,205}{modpack-robotics.github.io}}}
}

\begin{document}
\bstctlcite{IEEEexample:BSTcontrol}
\twocolumn[{%
\renewcommand\twocolumn[1][]{#1}%
\maketitle
\includegraphics[width=0.99\textwidth]{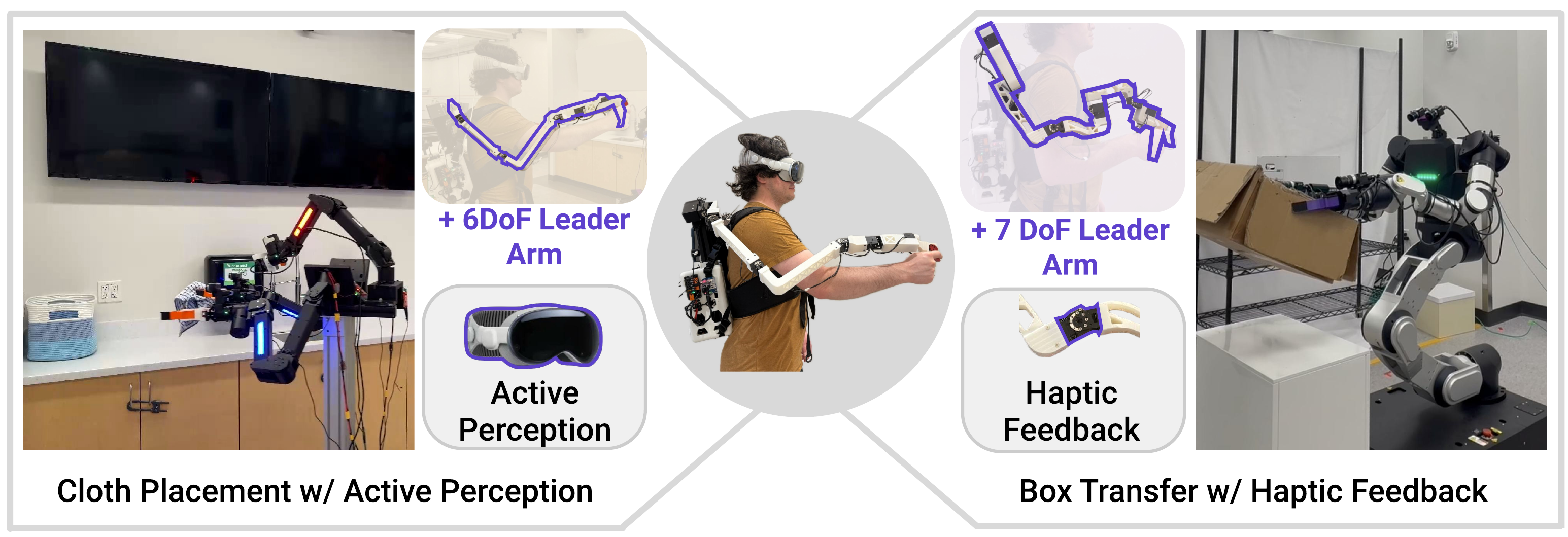}
\vspace{-2mm}
\captionof{figure}{\footnotesize \textbf{\ours} is an extensible teleoperation interface for bimanual mobile manipulation, built around a wearable backpack base with plug-and-play capability modules. For example, the active perception module, shown on the left, lets operators control the robot’s head camera to search for cloth placement locations, while the leader arms module with haptic feedback, shown on the right, supports a box transfer task by preventing excessive force and facilitating bimanual coordination. Swappable leader arms further allow adaptation to different robot embodiments.}
\vspace{3mm}
\label{fig:teaser}
}]
\thispagestyle{empty}
\pagestyle{empty}

%%%%%%%%%%%%%%%%%%%%%%%%%%%%%%%%%%%%%%%%%%%%%%%%%%%%%%%%%%%%%%%%%%%%%%%%%%%%%%%%
\begin{abstract}

Existing teleoperation systems are often tailored to specific robot hardware and task domains, limiting their scalability and adaptability. We present {ModPack}, a modular and extensible teleoperation system designed to support diverse robot embodiments and task requirements within a unified framework. At the core of ModPack is a self-contained wearable ``backpack'' that integrates onboard computation, power, communication, and data storage. Built on top of this shared interface, the system supports plug-and-play capability modules including joint-level teleoperation with haptic feedback, mobile manipulation, and active perception. Experiments across two distinct robot platforms and real-world mobile manipulation tasks demonstrate that ModPack provides a flexible and reusable framework for data collection and policy learning and collects higher quality data than a robot-free alternative. To support future research, we open-source the complete hardware design and software stack.

\end{abstract}

%%%%%%%%%%%%%%%%%%%%%%%%%%%%%%%%%%%%%%%%%%%%%%%%%%%%%%%%%%%%%%%%%%%%%%%%%%%%%%%%
\section{INTRODUCTION}

% \begin{figure*}[t]
%     \centering
%     \includegraphics[width=0.99\linewidth]{figures/teaser_v4.pdf}
%     % \vspace{-5mm}
%     \captionsetup{font=small}
%     \caption{\footnotesize \textbf{\ours} is an extensible teleoperation interface for bimanual mobile manipulation, built around a wearable backpack base with plug-and-play capability modules. For example, the active perception module, shown on the left, lets operators control the robot’s head camera to search for cloth placement locations, while the leader arms module with haptic feedback, shown on the right, supports a box transfer task by preventing excessive force and facilitating bimanual coordination. Swappable leader arms further allow adaptation to different robot embodiments.
%     % \todo{make figure bigger, change left task figure more active preception}
%     }
%     \label{fig:teaser}
%     \vspace{-3mm}
%     %https://www.figma.com/design/eiAH0RQlPKBrBsUeB9fEXt/Cover-Figure?node-id=29-23&t=vC3MeFdSGjeHzBnm-0
% \end{figure*}

Designing a teleoperation interface involves an inherent trade-off between system simplicity and functional capability. The most capable systems are often more complex and sometimes unnecessary for simpler applications. For instance, while active perception is vital for long-horizon search-and-reach tasks, it may introduce redundant overhead for tabletop manipulation. As a result, prior work has often required researchers to carefully and rigidly tailor teleoperation interfaces to match the requirements of a given robot hardware and task domain.

This task- and robot-specific specialization, while effective, leads to a fragmented design space. Systems optimized for one setting are often difficult to extend or adapt to others, limiting scalability and slowing iteration. In particular, adding new capabilities typically requires substantial redesign of both hardware and software pipelines. This lack of \textbf{modularity} presents a key bottleneck for building unified data collection and control systems that generalize across tasks and robot platforms.

In this work, we introduce \textbf{\ours} -- a teleoperation system that explicitly addresses this trade-off through an \textbf{extensible} and \textbf{modular} design. At its core, \ours~is a self-contained ``backpack" unit that provides integrated power, onboard computation, and data storage. This unit serves as a common substrate, decoupling core system infrastructure from task- or robot- specific functionalities. Built on this foundation, we propose a plug-and-play interface that allows operators to flexibly add or remove capability modules based on task requirements. We demonstrate the versatility of this design through several modular extensions, each enabling a distinct capability:

\begin{itemize}[leftmargin=6mm]

\item \textbf{Cross-robot joint-level control:} Through swappable leader-arm modules, as shown in \Cref{fig:teaser}, the system can adapt to diverse robot arm kinematics, enabling precise joint-level control across different robot platforms.

\item \textbf{Mobility:} We track the operator's base motion through an iPhone mounted on \ours, enabling mobile manipulation for tasks that require large workspace coverage.

\item \textbf{Active perception:} By wirelessly connecting to a VR headset, such as the Vision Pro, \ours~streams real-time robot head-camera observations while allowing the operator to control the camera pose. This module enables demonstrations of active perception behaviors, such as object search, occlusion handling, and viewpoint selection.

\item \textbf{Haptic feedback:} By leveraging force feedback from the robot and active motors on the leader arms, our system provides real-time haptic feedback, enabling safer and more precise teleoperation of contact-rich tasks.

\end{itemize}
% While these capabilities have been explored individually, integrating them into a unified, flexible, and robust teleoperation system is non-trivial. First, simultaneous real-time control of the arms, mobile base, and head requires \ours~to \textbf{disentangle operator motions} to component-wise control commands. Second, the system synchronizes \textbf{high-rate heterogeneous data}, including proprioception, visual observations, and force/torque for haptic feedback, while maintaining low latency and precise temporal alignment. 
% Third, extended operation requires an \textbf{unobtrusive and comfortable} interface, which we address with effective gravity compensation on the exoskeleton leader arms to reduce operator fatigue. 
% %
% Finally, these capabilities must fit within a \textbf{compact wearable form factor}; we package power, computation, sensing, and communication hardware into a self-contained backpack while navigating additional constraints on weight, thermal management, and usability.

\looseness-1
While these capabilities have been explored individually, integrating them into a unified, flexible, and robust teleoperation system remains challenging. First, simultaneous control of the arms, mobile base, and head introduces ambiguity in operator motion; \ours~addresses this by disentangling operator movements into component-wise commands for arm manipulation, base motion, and active perception. Second, the system must process high-rate heterogeneous data, including proprioception, visual observations, and force/torque signals for haptic feedback; we collect these streams with low-latency communications through message-queue servers. Third, extended operation requires an unobtrusive and comfortable interface; we reduce operator fatigue through gravity compensation on the exoskeleton leader arms. Finally, all capabilities must be integrated into a compact wearable form factor; we package power, computation, sensing, and communication hardware into a self-contained backpack while balancing constraints on weight, thermal management, and usability.

We evaluate \ours~through real-world bimanual mobile manipulation experiments across two distinct robot platforms. Using our teleoperation device, we collect demonstrations, train behavior cloning policies, and evaluate deployment success rates. To support future research, we will open-source the full hardware design, software stack, and policy learning and deployment framework.

\section{RELATED WORK}

\label{sec:related_works}
\mypara{Teleoperation for Robot Manipulation.} Teleoperation serves as a primary paradigm for collecting high-quality demonstration data for robot learning \cite{punamiya2026egoverse, barreiros2026careful, bjorck2025gr00t, team2025gemini, o2024open, khazatsky2024droid}. Early approaches such as kinesthetic teaching \cite{Argall2009LfD} enable direct physical guidance but are labor-intensive and difficult to scale. To improve usability, many systems decouple the operator from the robot, using input devices (e.g., spacemouse, joysticks, or VR controllers) to command end-effector poses with inverse kinematics ~\cite{Shivin_Wensi_Yuqian_Samik_Jiaheng_Ruohan_Peter_Ben_Roberto_2024, iyer2024open}. In contrast, kinematic replicas such as GELLO~\cite{wu2024gello} enable one-to-one joint mapping, improving control fidelity and granularity. As tasks move beyond static tabletop settings, teleoperation for mobile manipulation becomes critical. Mobile ALOHA~\cite{fu2024mobile} enables simultaneous base and bimanual arm control via a leader–follower setup, though base and arm control are mostly decoupled. TidyBot++~\cite{wu2024tidybot++} supports joint control of a mobile base and a single arm, but does not scale naturally to coordinated bimanual tasks.

To achieve more natural human-like control, recent works explore exoskeleton-based teleoperation with isomorphic mappings. AirExo and AirExo-2~\cite{fang2024airexo,fang2025airexo} provide high-fidelity arm control and cross-robot adaptability, but lack integrated mobile base control. Full-body systems such as CHILD~\cite{myers2025child} and PAPRLE~\cite{kwon2025paprle} support both arm and leg control, though typically not simultaneously. Most closely related, Supersuit~\cite{chen2026supersuit} and UME~\cite{liang2026universalmanipulationexoskeletonlearning} enable whole-body teleoperation with base and torso tracking, but neither provide both active perception and haptic feedback together. In contrast, ModPack covers key capabilities for mobile teleoperation, and in addition, allows users to select different combinations of components for their target system through its modular design. Table \ref{tab:capability_comparison} presents a detailed comparison with prior works.

% \josh{another work i found that we should include is \cite{Lenz_2023}, which has most of our features but relies on Panda arms and a foot pedal to steer}
% while HOMIE~\cite{ben2025homiehumanoidlocomanipulationisomorphic} explores similar ideas for humanoids.

\mypara{Teleoperation with Haptic Feedback.}
Haptic feedback is critical for collecting high quality demonstrations for contact-rich tasks. A key challenge in haptic feedback for teleoperation is balancing feedback fidelity with hardware portability. High-fidelity bilateral systems can provide rich force feedback, but often rely on stationary \cite{liu2025factrforceattendingcurriculumtraining} or off-the-shelf robot arms that are too heavy and cumbersome for wearable mobile interfaces~\cite{Lenz_2023,nimbro}. In contrast, lightweight wearable systems \cite{zhang2023wearable, purushottam2024wheeledhumanoidbilateralteleoperation, wu2025vrbased} and vibrotactile feedback \cite{ding2025bunny} improve portability, but typically sacrifice one-to-one joint-level transparency and kinesthetic resistance. In this work, we address this trade-off with a wearable teleoperation interface built from motorized GELLO arms integrated into a backpack-mounted frame. By actuating each joint directly, our system provides high-fidelity, joint-level haptic feedback while preserving the portability required for mobile teleoperation.

% Advanced wearable devices have attempted to offer joint-level feedback \cite{zhang2025doglove}, but often struggle with the interpretability of complex forces, leading some to explore task-space remapping \cite{Lenz_2023} or discrete contact categorization \cite{9991865}.

% Collaborative Teleop with Haptic Feedback \cite{9981426}: haptic feedback guidance to the path of robot to guide, human co-navigates autonomous robot

% Improving Teleoperation Through Human-Aware Haptic Feedback \cite{9991865}: creates different haptic feedback signals based on what type of contact robot is in, i.e. slipping versus contact

% Bimanual Telemanipulation with Force and Haptic Feedback \cite{Lenz_2023}: controller with haptic and force feedback, use an identical robot arm so a bit overkill, also has 6d head pose control with VR headset, mobile base controlled with 3D foot "rudder"

% Wheeled Humanoid Bilateral Teleoperation with Position-Force Control \cite{purushottam2024wheeledhumanoidbilateralteleoperation}: could also be relevant for backpack-esc design, haptic feedback through torso

% HoMMI \cite{xu2026hommilearningwholebodymobile}: Not teleop, UMI-style data collection, with active perception
% HoMeR \cite{sundaresan2025homerlearninginthewildmobile}: learns to predict relative vs absolute actions for end effector, uses TidyBot++ base, create a whole body controller
% EMMA \cite{zhu2026emma}: cotrain whole body human data with static teleoperation data

\begin{table}[t]
\label{sec:prior_table}
\centering
\footnotesize
\vspace{-2mm}
\setlength{\tabcolsep}{2pt}
\renewcommand{\arraystretch}{0.9}
\begin{tabular}{@{}lccccc@{}}
\toprule
 & \textbf{Cross-Robot} & \textbf{Joint-Level} & \textbf{Haptic} & \textbf{Mobile} & \textbf{Active} \\
 & \textbf{Adapt.} & \textbf{Control} & \textbf{Feedback} & \textbf{Manip.} & \textbf{Percep.} \\
\midrule
\cite{cheng2024open}  & \cmark & \xmark & \xmark & \xmark & \cmark \\
\cite{lu2025mobile}   & \cmark & \xmark & \xmark & \cmark & \cmark \\
\cite{wu2024gello}    & \cmark & \cmark & \xmark & \xmark & \xmark \\
\cite{fu2024mobile}   & \xmark & \cmark & \xmark & \cmark & \xmark \\
\cite{fang2025airexo} & \cmark & \cmark & \xmark & \cmark & \xmark \\
\cite{xiong2025vision}& \xmark & \cmark & \xmark & \xmark & \cmark \\
\cite{myers2025child, kwon2025paprle, chen2026supersuit} & \cmark & \cmark & \xmark & \cmark & \xmark \\
\cite{purushottam2024wheeledhumanoidbilateralteleoperation} & \xmark & \xmark & \cmark & \cmark & \xmark \\
\cite{zhang2023wearable, wu2025vrbased, ding2025bunny} & \cmark & \xmark & \cmark & \xmark & \xmark \\
\cite{liang2026universalmanipulationexoskeletonlearning} & \cmark & \cmark & \cmark & \cmark & \xmark \\
\cite{xu2026hommilearningwholebodymobile} & \cmark & \xmark & \cmark & \cmark & \cmark \\
\midrule
\textbf{\ours} & \cmark & \cmark & \cmark & \cmark & \cmark \\
\bottomrule
\end{tabular}
\vspace{1mm}
\caption{\footnotesize \textbf{Comparison} with existing data collection systems across key capabilities. \ours~supports the key capabilities studied in prior work within a unified, modular interface.}
\vspace{-5mm}
\label{tab:capability_comparison}
\end{table}

\mypara{Teleoperation with Active Perception. }
Active perception is the ability to adaptively change viewpoint to acquire task-relevant visual information. Recent works study active perception with head- or body-mounted sensing~\cite{xu2026hommilearningwholebodymobile, zeng2025activeumiroboticmanipulationactive, yu2025egomi}. In teleoperation, active perception is often achieved by tracking the operator's head pose with a VR headset and mapping it to robot camera motion while streaming the robot's egocentric view~\cite{cheng2024open, xiong2025vision}. However, existing systems primarily operate in stationary settings, where camera and base motion are decoupled by design. Mobile teleoperation introduces a key challenge: disentangling viewpoint changes caused by head motion from those caused by body or base motion. Our system tracks the operator’s head and body motion with the Vision Pro and a backpack-mounted iPhone, respectively, and compensates for base motion before sending adjusted robot head commands. This decoupling enables operators to naturally walk, look around, and control the arms during data collection.

\section{METHOD}
{\setlength{\textfloatsep}{0pt}
\begin{figure}[t]
    \centering
    \includegraphics[width=0.97\linewidth]{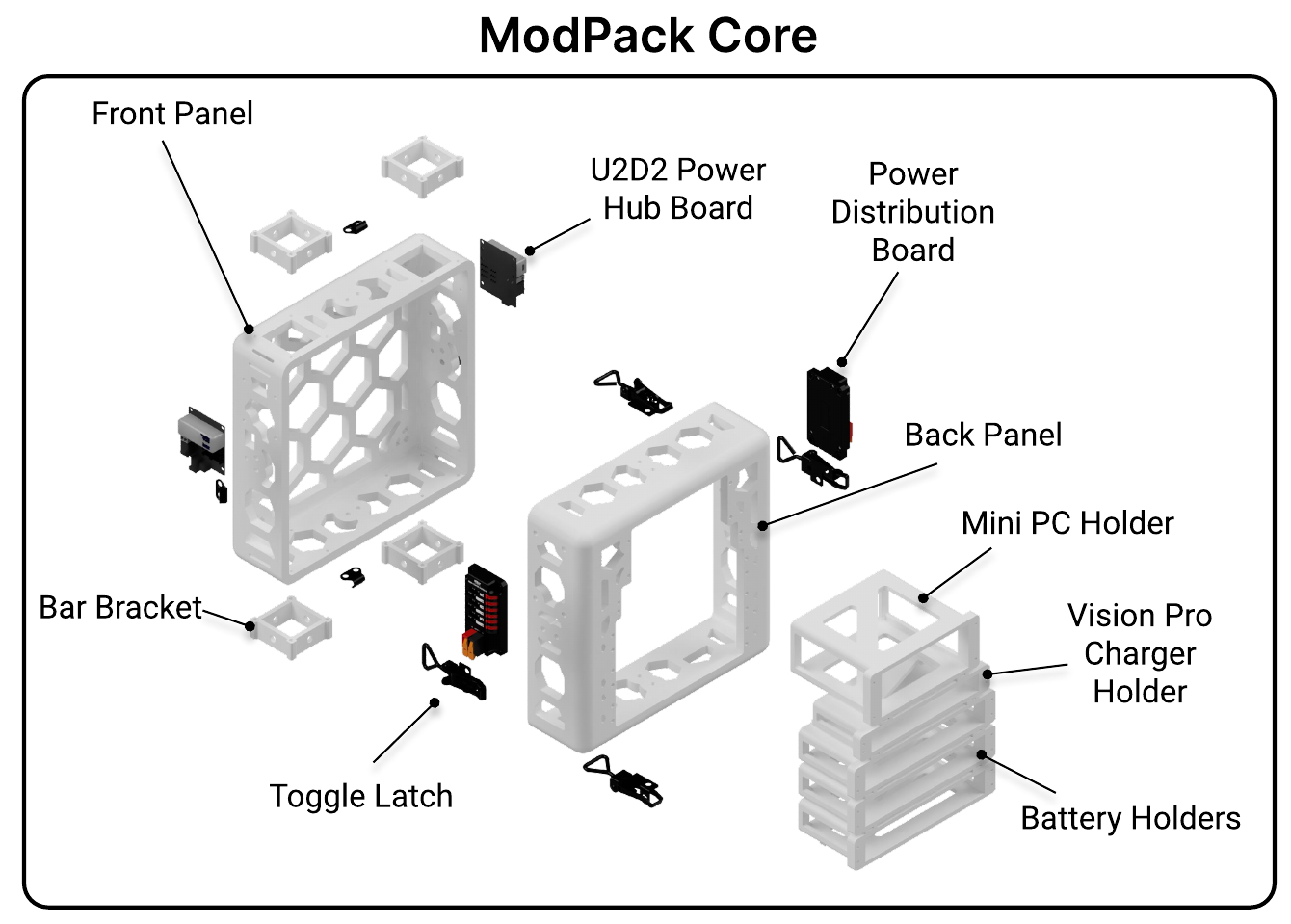}
    \vspace{-2mm}
    \caption{\footnotesize \textbf{\ours} consists of a 3D printed backpack core that supports plug-and-play modules, such as a Vision Pro for active perception and an iPhone for base control. The backpack contains onboard power as well as power distribution circuits using off the shelf components, powering the mini PC, leader arms, and optionally the Vision Pro charger.  
    % \todo{make caption longer, add more details}
    }
    \label{fig:method}
    %https://www.figma.com/design/eiAH0RQlPKBrBsUeB9fEXt/Cover-Figure?node-id=29-23&t=vC3MeFdSGjeHzBnm-0
    \vspace{-4mm}
\end{figure}}

\ours~is a self-contained, 3D-printed backpack system that provides onboard power, storage, and compute for mobile teleoperation. Its primary design goal is \textbf{extensibility}: users can easily add, remove, or replace hardware and software modules to support new robot embodiments with minimal setup. We describe the core system, implemented modules---including \textsc{leader arms}, \textsc{mobile base}, and \textsc{active perception}---and integrations with two distinct mobile robot platforms.

\subsection{\ours~Core}

At the core of \ours~is a portable backpack that serves as a shared base platform for modular hardware extensions and provides the software interfaces for integration. This section describes the core hardware and software architecture that enables these extensions.

\mypara{Hardware.}
The \ours~core is a wearable backpack base unit that moves with the operator while providing a configurable mounting and housing structure for modular hardware components. The backpack is fabricated from 3D-printed parts, improving accessibility and enabling rapid modification by the research community. As shown in Fig.~\ref{fig:method}, the main backpack assembly consists of 3D-printed front and rear panels secured with toggle latches, allowing convenient access to the internal components. The front panel includes two top-mounted slots for inserting support bars used to mount the leader-arm modules.

The internal volume is organized into five removable shelves that house the onboard compute, power, and optional accessories.
% Specifically, the backpack holds a Geekom A5 Mini PC, three 145,W, 25{,}000,mAh portable power banks, and an optional Vision Pro charger. Two power banks are assigned to the left and right leader arms, respectively, while the third powers the onboard PC.
Following ergonomic observations from prior backpack-based systems~\cite{genitrini2022backpacks}, which suggest that vertically distributed loads near the thoracic region can reduce carrying strain, the PC and power banks are arranged along the operator’s spine. To further reduce weight and improve thermal ventilation, excess material is removed from both the structural panels and internal shelves. 

\begin{figure}
    \centering
    \includegraphics[width=\linewidth, center]{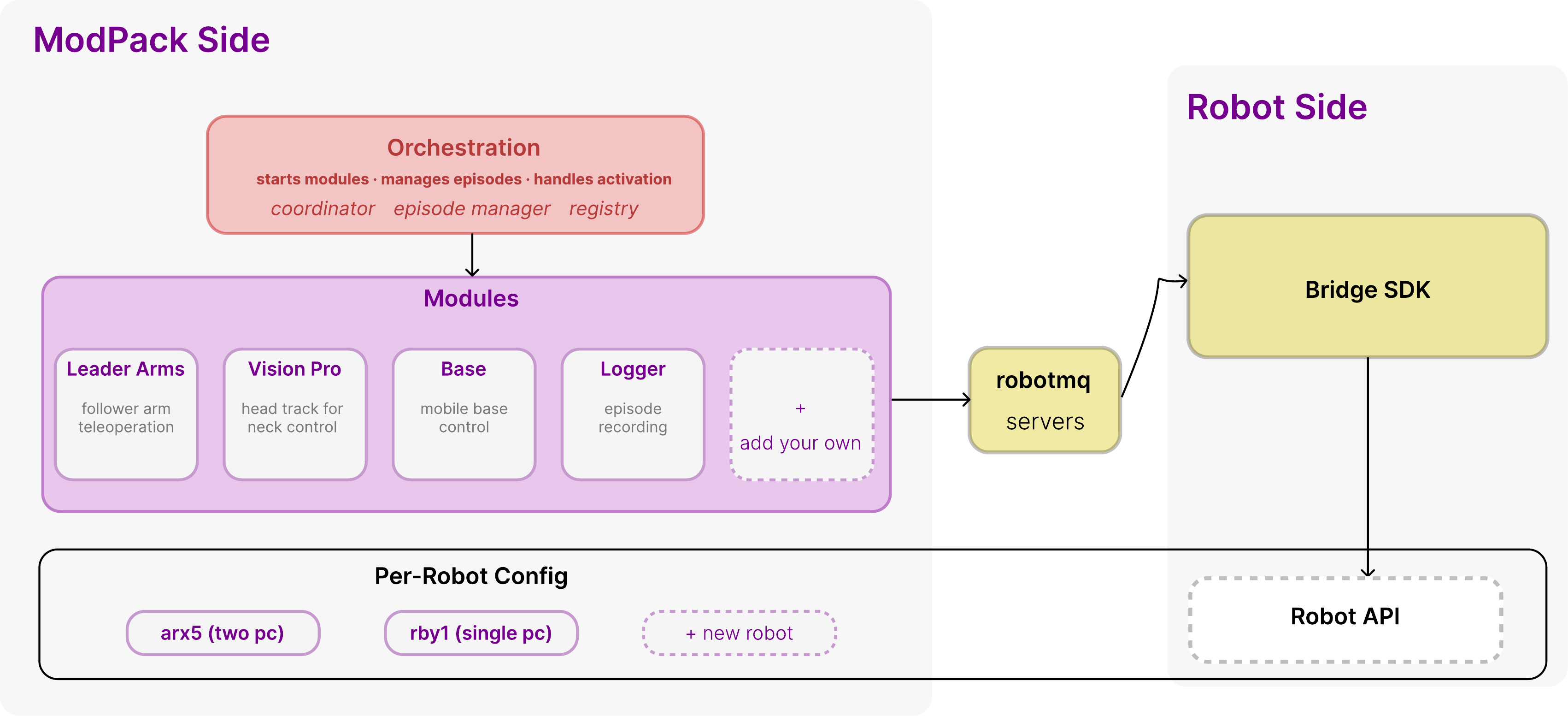}
    \caption{\footnotesize \textbf{Software diagram for ModPack. } Modules and data recording all occur on the mini PC within Modpack, while robot specific code is handled through a lightweight bi-directional bridge API.}
    \label{fig:software}
    \vspace{-6mm}
\end{figure}

\label{sec:software}
\mypara{Software.} The onboard mini PC serves as a standalone compute unit for module activation, data logging, and target-robot startup and shutdown. Target robots communicate with the backpack PC through a lightweight bidirectional ``bridge'' API. The underlying communication is abstracted from the user and implemented using a lightweight message queue, following~\cite{gao2026gatedmemorypolicy}. This abstraction decouples the core backpack stack from embodiment-specific implementations, allowing the same system to support different robots and arbitrary subsets of extended modules. A diagram of the software stack can be seen in Figure \ref{fig:software}.

% \section{Software Diagram}
% Software is split into two sides as seen in Figure~\ref{fig:software}. \ours~handles lifecycle management and modules while setting relevant configurations for downstream robots. The per-robot configurations define the bridge topics and what modules are turned on, meaning that all the end user must do is call the relevant bridge API for bi-directional communication with \ours~through \texttt{robot-message-queue} \cite{gao2026gatedmemorypolicy}. Actions flow from module commands to robot through \texttt{robot-message-queue} servers. Data is saved in a \texttt{zarr} format using \texttt{robologger}, an open source logging pipeline, which coordinates individual loggers with a main logger.

% To support extensibility, we define a standardized workflow for adding new modules. Each module is implemented as an independent process, registered through a common interface, and optionally managed through a shared lifecycle controller. This design allows users to easily extend \ours~with new hardware or software capabilities while preserving a consistent system interface.

\subsection{Extensible Modules}
The core system is extensible through modular components, each consisting of a hardware unit and its corresponding software interface. This design enables physical modules to be swapped across target robots while preserving a unified API and minimizing code changes. Below, we describe three modules---\textsc{leader arms}, \textsc{mobile base}, and \textsc{active perception}.

\subsubsection{Leader Arms}
\label{method:leader-arms}
% \mypara{Hardware design.}
% additional fig specifically for 1-1 config example? flow from each cad limb to each robot arm, motor to motor arrows in diagram
As shown in Fig.~\ref{fig:method}, we design and 3D-print two leader-arm variants: a pair of 6-DoF ARX5 leader arms for a customized mobile manipulation platform \cite{xiong2025vision}, and a pair of 7-DoF leader arms for the RB-Y1m robot. We used Dynamixel motors sized to provide the minimum amount of torque necessary to hold the arm for mass assumptions made using OnShape materials library. For RB-Y1m leader arms, we ensured a high enough maximum current limit to support haptic feedback. Details on current limits and motor selection can be found in Table \ref{tab:leader_arms}.

\mypara{Customized mobile robot \cite{xiong2025vision}.} Our customized mobile robot uses bimanual 6-DoF ARX5 arms and a 6-DoF ARX5 for active perception as presented in \cite{xiong2025vision}, mounted on a four-wheeled Tidybot++ base from \cite{wu2024tidybot++}. The neck is equipped with an iPhone that records depth and RBG images; the two arms have Logitech webcams to capture wrist RBG images. Capture rates are detailed in Section \ref{sec:policy}. We leave joints 5 and 6 unactuated to allow for better wrist mobility. Current limits and damping gains were tuned empirically for stable gravity compensation with minimal stiffness and jitter, with $k_d = 0.1$ for all motors. 

% Each joint motor is clipped to an empirically determined current limit, which was chosen to provide stable gravity compensation without making the joint too stiff. Damping gains were chosen empirically to reduce jitter.
\begin{table}[htbp]
\centering
\small
\begin{tabular}{cclc}
\toprule
Arm & Joint & Motor Model & Current Limit (mA) \\
\midrule
\multirow{7}{*}{6-DoF}
 & 1       & XM540-W270-T & 100 \\
 & 2       & XM540-W270-T & 200 \\
 & 3       & XM430-W350-T & 200 \\
 & 4       & XM430-W350-T & 60  \\
 & 5       & XL430-W250-T & --- \\
 & 6       & XL430-W250-T & --- \\
 & Gripper & XL330-M288-T & --- \\
\midrule
\multirow{8}{*}{7-DoF}
 & 1       & XM540-W270-T & 200 / 150 \\
 & 2       & XM540-W270-T & 50 / 50   \\
 & 3       & XM540-W270-T & 100 / 100 \\
 & 4       & XM540-W270-T & 100 / 100 \\
 & 5       & XM540-W270-T & ---       \\
 & 6       & XM430-W350-T & 100 / 80  \\
 & 7       & XM430-W350-T & ---       \\
 & Gripper & XL330-M288-T & ---       \\
\bottomrule
\end{tabular}
\caption{\footnotesize \textbf{Motor specifications and control parameters} for the 6-DoF and 7-DoF leader arms. The 7-DoF arms lists gravity / contact for current limits.}
\label{tab:leader_arms}
\vspace{-4mm}
\end{table}
% \todo{cite via and how this robot differs, camera configs}
\mypara{RB-Y1m.}
The RB-Y1m robot is a bimanual setup with 7-DoF arms and a 6-DoF actuated torso on a holonomic mobile base. Using the setup from \cite{xu2026hommilearningwholebodymobile}, we use a 2-DoF active neck as well as fin-ray fingers as grippers. Continuing with the setup from \cite{xu2026hommilearningwholebodymobile}, we use 2 wide-angle cameras, \texttt{FLIR
BFS-PGE-23S3C-CS}, and one \texttt{FLIR BFS-PGE-50S5C-C} on each wrist of the two 7-DoF arms. For the associated leader arm, we choose to not actuate joints 5 and 7 to allow for more stable control and better wrist mobility, respectively. Damping gains are zero for all motors.

% \begin{table}
% \centering
% \begin{tabular}{clcc}
% \toprule
% Joint & Motor Model & \multicolumn{2}{c}{Current Limit (mA)} \\
% \cmidrule(lr){3-4}
%       &              & Gravity & Contact \\
% \midrule
% 1       & XM540-W270-T & 200 & 150 \\
% 2       & XM540-W270-T & 50  & 50  \\
% 3       & XM540-W270-T & 100 & 100 \\
% 4       & XM540-W270-T & 100 & 100 \\
% 5       & XM540-W270-T & --- & --- \\
% 6       & XM430-W350-T & 100 & 80  \\
% 7       & XM430-W350-T & --- & --- \\
% Gripper & XL330-M288-T & --- & --- \\
% \bottomrule
% \end{tabular}
% \caption{Motor specifications and control parameters for 7-DoF leader arm.}
% \label{tab:rby1_leader_arm}
% \end{table}
Given the CAD model of a target robot, each leader arm is constructed to be kinematically equivalent to its corresponding follower arm, following GELLO~\cite{wu2024gello}, enabling direct joint-space teleoperation. 

\mypara{Gravity Compensation.} To alleviate operator fatigue during teleoperation, we implement active gravity compensation \cite{shi2026minimalistcompliancecontrol} on the leader arms. The gravity torque vector $\tau_g = G(q)$ is computed using the Orocos Kinematics and Dynamics Library (KDL) in conjunction with the leader arm's URDF model. To enhance stability, we optionally append a joint-space damping term to form the augmented torque vector at time $t$: $\tau'_t = \tau_{g,t} + K_d \dot{q}_t$, where $K_d$ is a diagonal damping gain matrix. To ensure smooth and stable assistance, the augmented torque for each joint $i$, denoted as $\tau'_{i,t}$, is filtered using two cascaded exponential moving average (EMA) filters:
\begin{equation}
    \tau_{i,t}^{\text{inter}} = \alpha \tau_{i,t-1}^{\text{inter}} + (1-\alpha)\tau'_{i,t}, 
    \quad
    \tau_{i,t}^{\text{final}} = \alpha \tau_{i,t-1}^{\text{final}} + (1-\alpha)\tau_{i,t}^{\text{inter}},
\end{equation}
where $\alpha \in [0, 1)$ is the smoothing factor. The final torque vector is scaled by a gain factor and commanded to the leader arm motors, subject to the maximum current constraints configured.

% \maxcomment{Text is too small in Figure 3}
\begin{figure}
    \centering
    % \vspace{-mm}
    \includegraphics[width=0.98\linewidth]{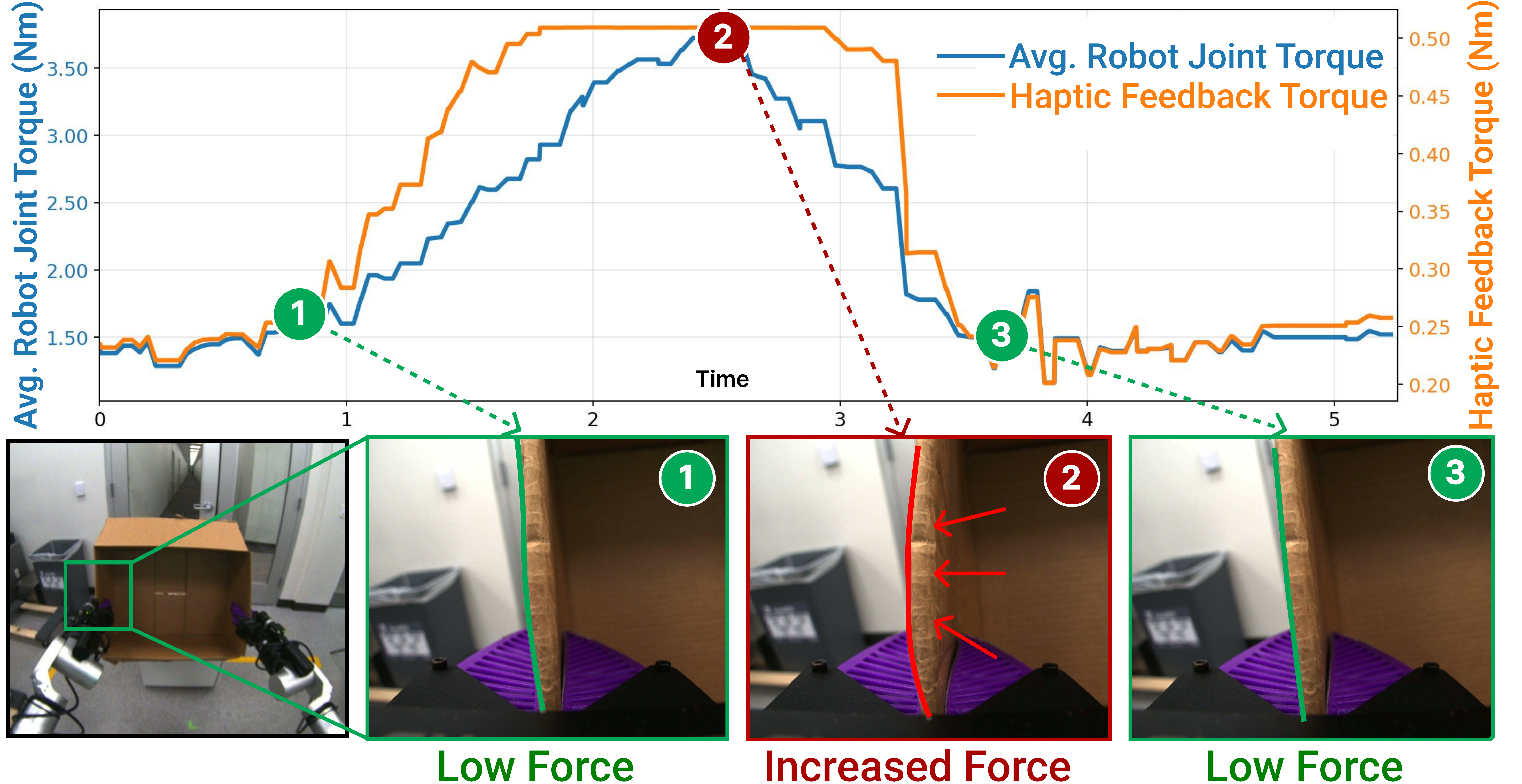}
    \caption{\footnotesize \textbf{Haptic feedback.} The top plot shows the average joint torque on RB-Y1m left arm (blue), along with the corresponding haptic feedback torque norm commanded to the left leader arm (orange). The bottom panel shows operator’s response to the haptic feedback to adjust the force exerted on the box.
    }
    \label{fig:haptic}
    \vspace{-6mm}
    %https://www.figma.com/design/eiAH0RQlPKBrBsUeB9fEXt/Cover-Figure?node-id=29-23&t=vC3MeFdSGjeHzBnm-0
\end{figure} 

\mypara{Haptic Feedback.} To enhance operator situational awareness, we implement haptic feedback for follower robots equipped with force/torque (F/T) sensors, in our case, the RB-Y1m robot. 
When enabled, the system reads the net wrench at the end-effector, $F_{ee}$, and maps it to joint-space torques via the Jacobian transpose: $\tau_{c} = k_{c} \cdot J^TF_{ee},$
% \begin{equation}
%     \begin{aligned}
%         \tau_{c} &= k_{c} \cdot J^TF_{ee},\\
%     \end{aligned}
% \end{equation}
where $k_{c}$ is a tuneable gain. 
The final control torque $\tau_{f}$ commanded to the leader motors is the superposition of the gravity compensation torque and the haptic feedback component: $\tau_{f} = \tau_{g_i,t}^\text{final} + \tau_{c}$.
% \begin{equation}
%     \tau_{f} = \tau_{g_i,t}^\text{filtered} - \tau_{c}.
% \end{equation}
To maintain hardware longevity and accommodate lower-cost actuators, we implement a current budgeting strategy. This limits peak torque commands and ensures the motors operate within conservative amperage constraints, facilitating safe operation without compromising the transparency of the force feedback.

% This formulation allowing users to perceive contact interactions while maintaining smooth and stable teleoperation, ensuring safe operation and consistent feedback across a range of actuator capabilities.

\subsubsection{Mobile Base}
% \todo{@josh take a pass, I'm a bit confused by the notation. what is p? is it the 6DoF pose, or 3D position? what is R, is it the rotation matrix?}\josh{p is position vector (x,y,z) and R is the rotation matrix yes}
We track the operator’s $SE(2)$ pose using a backpack-mounted iPhone, which runs a modified WebXR session following \cite{wu2024tidybot++}. To ensure intuitive base control, we implement an egocentric mapping procedure that couples the robot’s base translation directly to the operator’s displacement within their local frame. For example, when the operator walks forward, the robot moves forward in its own local frame.

To recover the operator's pose in the robot frame, we first transform the raw pose from WebXR $(p_{xr}, R_{xr})$ to align with the robot's coordinate convention and obtain $(p_{r}, R_{r})$. The key challenge is then that the operator and iPhone rotate about different centers, as the iPhone is mounted to the backpack and not the operator. To account for the physical displacement between the iPhone and the operator, we apply sequential offsets to recover the operator’s center of rotation $p_o$: a device-specific offset $p_d$ to find the camera mount center $p_c$ \cite{wu2024tidybot++}, then an operator-specific offset $p_u$ to recover the user's true center of rotation $p_o$, where $R_c$ is the camera mounting orientation:
\begin{equation}
p_{c} = p_{r} + R_{r}p_{d}, \quad p_{o} = p_{c} + R_{c}p_{u}
\end{equation}
% This sequential transformation shifts the tracking reference from the physical sensor to the operator's body center. Consequently, the resulting trajectory represents the user's true translational intent rather than the arc-like swinging motion typical of a backpack-mounted sensor during body rotations.
This transformation ensures that the tracked trajectory represents the user's actual body movement rather than the motion of the backpack-mounted sensor.

To synchronize the coordinate systems, we map the operator’s tracking pose $(p_o, R_o)$ from the WebXR session frame into the robot’s world frame. Given the session's initial pose $(p_s, R_s)$ and the robot's global starting pose $(p_w, R_w)$, the world-relative target position $p_t$ is computed via:
\begin{equation}
R_{\text{rel}} = R_w R_s^{-1}, \quad p_t = p_w + R_{\text{rel}}(p_o - p_s)
\end{equation}
The target orientation matrix is similarly updated as $R_t = R_{\text{rel}} R_o$. From the resulting world-relative target state $(p_t, R_t)$, we extract the planar $x, y$ coordinates and the yaw heading, which are then issued as absolute target setpoints to the robot base controller.

% It is important to note that the iPhone and the center of rotation of the operator are fundamentally different. Thus, to recover the operator's pose in the robot frame, we apply a series of transformations. Because we are interested in the operator's pose and frame and not that of the iPhone, we must apply a set of transformations. Firstly, we apply a change of basis to the robot convention. This can be expressed as 
% \begin{equation}
%     \begin{aligned}
%         p_{r} &= A\,p_{xr}\\
%         R_{r} &= A \, R_{xr} \, A^T,
%     \end{aligned}
% \end{equation}
% where $A$ maps the WebXR frame to the robot basis and $p_{xr}, R_{xr}$ are the position and rotation of the iPhone in the WebXR frame, respectively. We then apply a device offset $p_{d}$ to get center of rotation as done in \cite{wu2024tidybot++}:
% \begin{equation}
%     p_{c} = p_{r} + R_{r}\,p_{d}.
% \end{equation}
% Next, we apply another offset $p_u$ to account for the difference in centers between the iPhone and the operator expressed as $p_{o} = p_{c} + R_{c}\,p_{u}$. 
% % \begin{equation}
% %     p_{o} = p_{c} + R_{c}\,p_{u}.
% % \end{equation}
% We convert the pose to a delta in the starting position $p_w$ and rotation $R_{w}$ for the WebXR session, which may be different from the current episode's starting position $p_s$ and rotation $R_s$.
% \begin{equation}
%     \begin{aligned}
%         R_{rel} &= R_w R_s\\
%         p_{t} &= p_w + R_{rel}\,(p_o - p_s).
%     \end{aligned}
% \end{equation}
% Finally, we take the $x,y$ and yaw values and convert them to absolute commands to send them as targets in base frame.

\subsubsection{Active Perception}

We implement the active perception module with an Apple Vision Pro running a custom visionOS app, which streams the operator’s head pose as robot neck commands similar to~\cite{xiong2025vision}. The interface provides a low-latency egocentric view of the robot, rendering either monocular RGB video or 3D point clouds in real time.

To map the operator's head motion, we track the current device pose $T_c \in SE(3)$ relative to the initial session pose $T_s \in SE(3)$, yielding the relative transformation $T_{\text{rel}} = T_s^{-1}T_c$. We then convert this motion into the robot head coordinate convention using a fixed basis transformation $B \in SO(3)$, mapping the relative translation and rotation as $p_r = Bp_{\text{rel}}$ and $R_r = BR_{\text{rel}}B^\top$. Together, these define the robot-frame head command $T_r \in SE(3)$. Finally, we transform $T_r$ into the robot world frame using the initial tracking reference $T_{\text{init}}$, yielding the global target head pose $T_{\text{target}} = T_{\text{init}}T_r$.

% \maxcomment{Might be good to preempt coment about this not being a problem if there's a unified coordinate system, could be shorter note.}

\mypara{Decoupling Head and Base Motion.}
We cannot directly command $T_{\text{target}}$, as it is expressed in the robot world frame and includes motion already induced by the mobile base. To avoid redundant commands during simultaneous navigation and active perception, we compensate for the current base pose $T_{\text{base}} \in SE(3)$ and express the target in the local robot base frame:
$T_{\text{neck}} = T_{\text{base}}^{-1} T_{\text{target}}$.
This isolates the residual neck motion intended by the operator.

% \mypara{Decoupling Head and Base Motion.} When an operator yaws their whole body, the neck and base both record rotation separately. This means that the neck will over-rotate. The same issue occurs for position. To prevent this, we must account for base movement before sending target poses to the neck. We first transform the relative commanded pose $T_r$ into the neck frame using the neck start pose, yielding $T_{er} = T_a T_r$. We then compensate for the current base motion using the base transform $T_b$, giving $T_{ee} = T_b^{-1} T_{er}$.
% We transform the relative commanded pose $T_r$ from the base frame to the neck frame $T_{er}$ using the base position $T_b$ and neck start pose $T_a$. We then compensate for the base movement using $T_b$. Putting this together, we compute this as
% \begin{equation}
%     \begin{aligned}
%         T_{er} &= T_a\,T_r\\
%         T_{ee} &= T_b^{-1}\,T_{er}.
%     \end{aligned}
% \end{equation}

\subsection{Policy Learning}
\label{sec:policy}
We train a standard transformer-based Diffusion Policy model \cite{chi2023diffusion} for each task using the collected teleoperation demonstrations. For the customized mobile robot, the policy takes as input RGB-D observations from the head camera and RGB observations from the left and right wrist cameras. For the RB-Y1m robot, the policy takes as input RGB observations from two head cameras and the left and right wrist cameras, and optionally left and right arm joint torques.
\looseness-1
All visual inputs are encoded using a CLIP-pretrained ViT, with depth images repeated across three channels before encoding. Torque observations are encoded with sequential CausalConv layers following~\cite{lee2019icra}. The resulting observation tokens are concatenated and cross-attended to the DiT model as conditioning inputs.

During training, we synchronize all sensory inputs to the downsampled head-camera timestamps (e.g., 10-20 Hz). Proprioceptive states are interpolated at each head-camera timestamp, with positions and gripper states interpolated linearly and orientations interpolated using SLERP. Other modalities are aligned by selecting the nearest preceding measurement to the head-camera timestamp.

For the custom mobile robot, the action space consists of a 3-DoF mobile-base command $(x, y, \theta)$, a 6-DoF head-camera end-effector pose, and bimanual arm commands. We represent the head pose using 3D translation and the 6D rotation representation, yielding a 9-dimensional head action. Each 6-DoF arm is controlled by six joint positions and a one-dimensional gripper position, resulting in 7 dimensions per arm. Thus the full action dimension is $26$. We predict base commands relative to the current base state, while the head, arm, and gripper commands are represented as absolute targets.

For the RB-Y1m robot, the action space consists of a 3-DoF mobile-base command $(x, y, \theta)$, a 2-DoF neck command corresponding to pitch and yaw, and bimanual arm commands. Each 7-DoF arm is controlled by seven joint positions and a one-dimensional gripper position, yielding 8 dimensions per arm. The full action dimension is therefore $21$. As with the custom mobile robot, we predict the base command relative to the current base state, while the neck, arm, and gripper commands are represented as absolute targets.

During inference, the model denoises an action chunk of length 16 using DDIM with 16 inference steps. We execute the first 8 actions for the cloth placement task and the full 16-action chunk for the box transfer task.

\begin{figure*}[t]
    \centering
    \includegraphics[width=\linewidth, center]{figures/task_1_v4.pdf}
    \caption{\footnotesize \textbf{Summary of Cloth Placement task.} a) The robot picks up a cloth, uses active perception to locate a basket placed on either side, navigates toward it, and places the cloth inside. b) Different evaluation scenarios with varying towel colors and basket locations. c) Task success rates across policy variants over input visual modalities. d) Failure cases across different policies.
    }
    \label{fig:results-task1}
\vspace{-4mm}
\end{figure*}
% https://www.figma.com/design/eiAH0RQlPKBrBsUeB9fEXt/ModPack-Figure?node-id=449-16

\section{EXPERIMENTS}
% \maxcomment{Small note: maybe use texttt or textsc instead of italic for the module names to make it stand out more}

To demonstrate \ours's adaptibility to different robot embodiments, we evaluate \texttt{Cloth Placement with Active Perception} task on the customized bimanual robot with ARX5 arms from \cite{xiong2025vision} with added holonomic base \cite{wu2024tidybot++}, and the \texttt{Box Transfer with Haptic Feedback} on an RB-Y1m robot. We also compare policy performance to HoMMI \cite{xu2026hommilearningwholebodymobile}, a robot free data collection and policy learning framework which uses two hand-held grippers and a head-mounted camera, for the box transfer task. We use this as a point of comparison for evaluating whether the added complexity of a full teleoperation system is justified and to directly test whether faster, cheaper data collection can substitute for embodiment-matched demos, as its capabilities are the closest to our system. During collection, \ours~was configured to use the appropriate \textsc{leader arms} module, the \textsc{mobile base} module, and the \textsc{active perception} module. For the RB-Y1m robot, we also utilize haptic feedback on the leader arm. We discuss the tasks below, followed by a summary of results.

\subsection{Cloth Placement with Active Perception}

\textbf{Task Description. } As shown in Figure \ref{fig:results-task1} a) and b): a cloth is placed in front of the robot on a counter, where a basket is either to the left or to the right of the robot. The robot must move forward to grasp the cloth and then reverse, at which point it should locate the basket on either side and traverse towards it, finishing the task once it places the cloth in the basket. We collect data with two cloths (blue, pink), two basket locations (left, right), and randomize initial positions for the cloth. The task is considered a success if \textbf{the cloth is dropped into the basket}. We collected 125 demonstrations in total for this task.
% This task highlights four distinct skills: \emph{Whole-system coordination}: The robot must control the base, neck, and arm movements simultaneously throughout the entirety of the task. \emph{Active perception}: The robot moves its neck to locate the basket. \emph{Navigation}: The robot must accurately move its base towards the basket and orient appropriately to drop the cloth. \emph{Bimanual coordination}: Both arms are actively controlled, with the right arm performing manipulation while the left arm maintains a steady position.
% For each scenario, we perform 5 rollouts, for 20 total evaluation trials. We additionally evaluate 5 challenge trials: one rollout from each scenario and one fully randomized trial varying basket side, cloth color, robot pose, and cloth pose, for 25 total trials overall.

\textbf{Results.}
We evaluate the task across scenarios with varying basket locations (left, right) and cloth colors (blue, pink). We compare three policies with different visual observation modalities: head camera only [Head Cam], wrist cameras only [Wrist Cam], and all cameras (head + both wrist cameras) [All Cam]. For each policy, we perform 25 rollouts with randomized cloth initial positions. 
% See a breakdown of the evaluation scenarios in Appendix \ref{sec:cloth_conditions}.
Quantitative results are shown in \Cref{fig:results-task1} c). [Head Cam] policy achieves the highest success rate, succeeding in 22/25 rollouts. [All Cam] policy succeeds in 20/25 rollouts, while [Wrist Cam] policy succeeds on only 3/25 rollouts. We hypothesize that the head camera performs best because the cloth admits many viable grasps, reducing the need for close-up wrist observations, while the head camera provides depth and a broader egocentric view for grasping and basket localization. Moreover, using only the head camera avoids potential out of distribution wrist camera observations caused by variations in wrist pose and cloth deformation. In contrast, [Wrist Cam] most often fails due to placement errors, as the policy lacks sufficient context of the basket location. Figure~\ref{fig:results-task1} d) shows a common failure mode for the [Wrist Cam] policy, which often dropped the cloth on the edge or out of the basket.

\begin{figure*}
    \centering
    \includegraphics[width=\linewidth, center]{figures/task_2_v4.pdf}
    \caption{\footnotesize \textbf{Summary of Box Transfer task.} a) The robot picks up a box, rotates to observe the existing box on the shelf and infer the target placement location, and then places the box accordingly. b) Different task scenarios with varying shelf availability configurations. c) Task success rates across policy variants, comparing the full policy against head-camera-only and wrist-camera-only ablations. d) Failure cases across different policies.}
    \label{fig:results-task2}
\vspace{-6mm}
\end{figure*}
% https://www.figma.com/design/eiAH0RQlPKBrBsUeB9fEXt/ModPack-Figure?node-id=724-2

\subsection{Box Transfer with Haptic Feedback}
\label{subsec:box-transfer}
\textbf{Task Description.} As shown in Figure \ref{fig:results-task2} a) and b), the robot is required to pick up a box from a cabinet, move to the rack, and place the box onto the unoccupied shelf, which can be either at the top or bottom. A trial is considered successful if \textbf{the robot places the box in a stable pose on the unoccupied shelf}. We collected 102 demonstrations in total, half for each of the two scenarios.

\begin{figure}
 %\vspace{-2mm}
    \centering
\includegraphics[width=0.4 \textwidth]{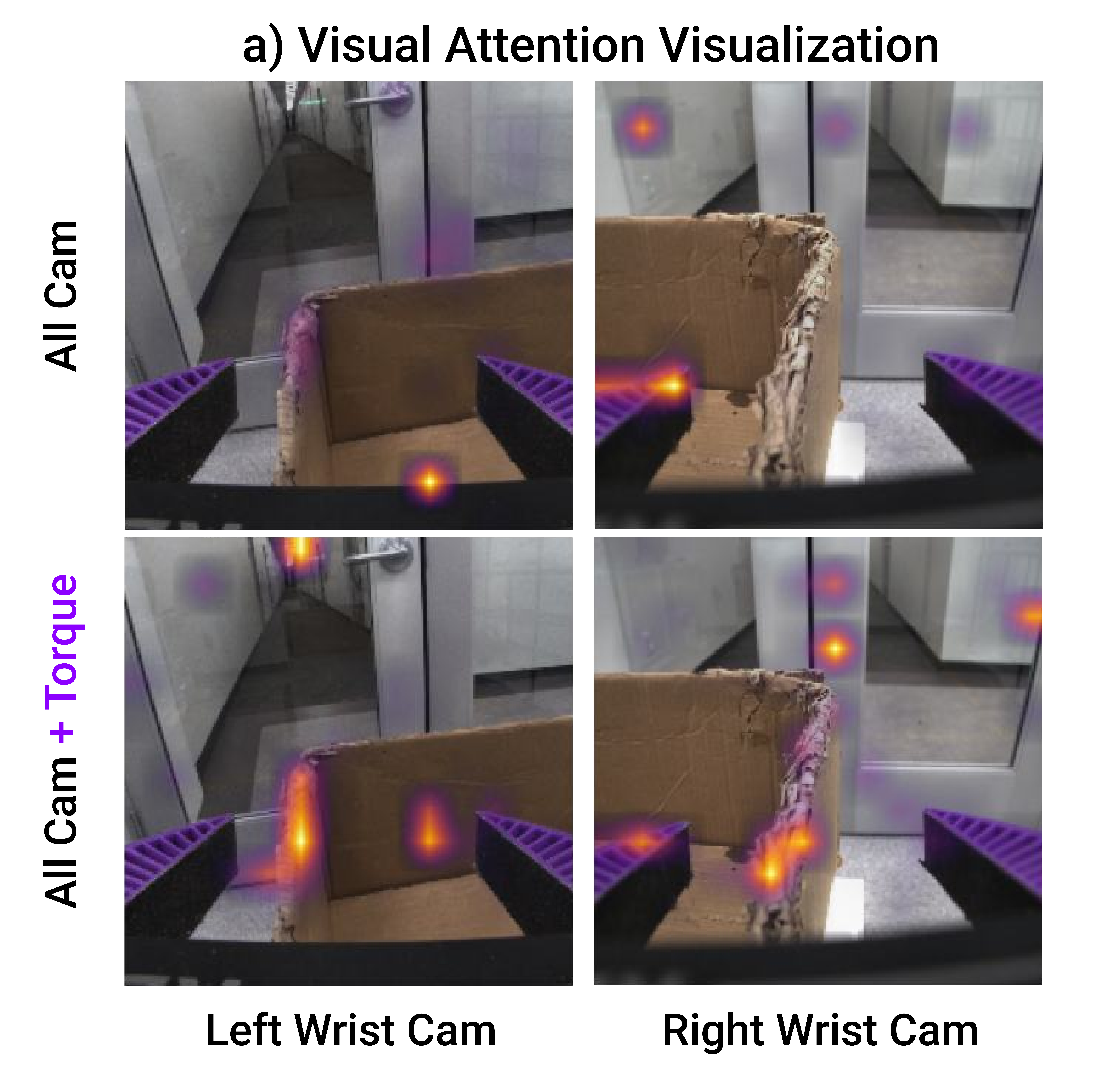}
    \caption{\footnotesize \textbf{Attention Analysis. } The All Cam policy's visual attention is highest towards the inside of the box on the left wrist, while the policy's attention on the right wrist is high on the gripper tip. All Cam + Torque policy's attention for both cameras is higher in task-relevant areas such as the sides of the box.}
    \label{fig:attention}
\vspace{-4mm}
\end{figure}

% This task displays four capabilities: \emph{Whole-system coordination}: The robot must continue to grasp the box during transport. \emph{Bimanual coordination}: The robot must use both arms to grasp the box. \emph{Active perception}: The robot must keep the box in view for grasp adjustments. \emph{Navigation}: The robot must traverse to the placement location to place the box.

% \begin{itemize}[leftmargin=*, topsep=0pt, itemsep=0pt, parsep=0pt, partopsep=0pt]
%     \item \textbf{Initial stage:} The robot approaches the table and grasps the box using both arms. The goal location is not in view.
%     \item \textbf{Intermediate stage:} The robot lifts and transports the box while searching across a large workspace toward the goal location. Once the goal location is in view, it navigates there. 
%     \item \textbf{Goal stage:} The robot aligns with the target and places the box precisely. 
% \end{itemize}

\textbf{Results. }
We evaluate the task across two scenarios where either the top or bottom shelf is occupied and the robot has to place the box on the other unoccupied shelf. We compare three policy variants: one using all RGB camera views (two head cameras and both wrist cameras) [All Cam], one using only the two head cameras [Head Cam], and one that takes in all camera views plus joint torque information for both arms [All Cam + Torque]. For each policy, we conduct 20 rollouts across two evaluation scenarios, varying the cart location where the box is initially placed and the robot’s starting positions.
% A detailed breakdown of the evaluation scenarios is provided in Appendix~\ref{sec:bt_conditions}. \josh{TODO: remove appendix references or add to main text}

Quantitative results are shown in Fig.~\ref{fig:results-task2} c). The policy conditioned on all camera views and joint torques [All Cam + Torque] achieves the highest success rate, succeeding in 12/20 rollouts, while [Head Cam] policy succeeds in 11/20 rollouts, and [All Cam] policy succeeds in 6/20 rollouts. As seen in Figure~\ref{fig:results-task2} d), a common failure mode of the [All Cam] policy is bad grasp poses. We hypothesize that incorporating projected joint torques improves performance by providing a coarse alignment signal for grasp timing and depth.

We provide a qualitative analysis of the model’s attention patterns. As shown in \Cref{fig:attention} a), we visualize the attention weights over image patches from the final layer of the ViT encoder for both the [All Cam] and [All Cam + Torque] policies. The torque-conditioned policy appears to attend more  to task-relevant regions, such as the edges of the box that are informative for grasping, while the vision-only policy attends to less relevant regions. In \Cref{fig:attention} b), we visualize the aggregated attention weights assigned to the torque tokens in the DiT model for the [All Cam + Torque] policy. Attention weights for both the left- and right-arm torque tokens increase as the robot approaches the box and initiates grasping, suggesting that the policy increasingly relies on torque information during contact-critical phases of the task.

\begin{figure}[t]
\label{fig:hommi}
% \vspace{-6mm}
    \centering

\includegraphics[width=0.47\textwidth]{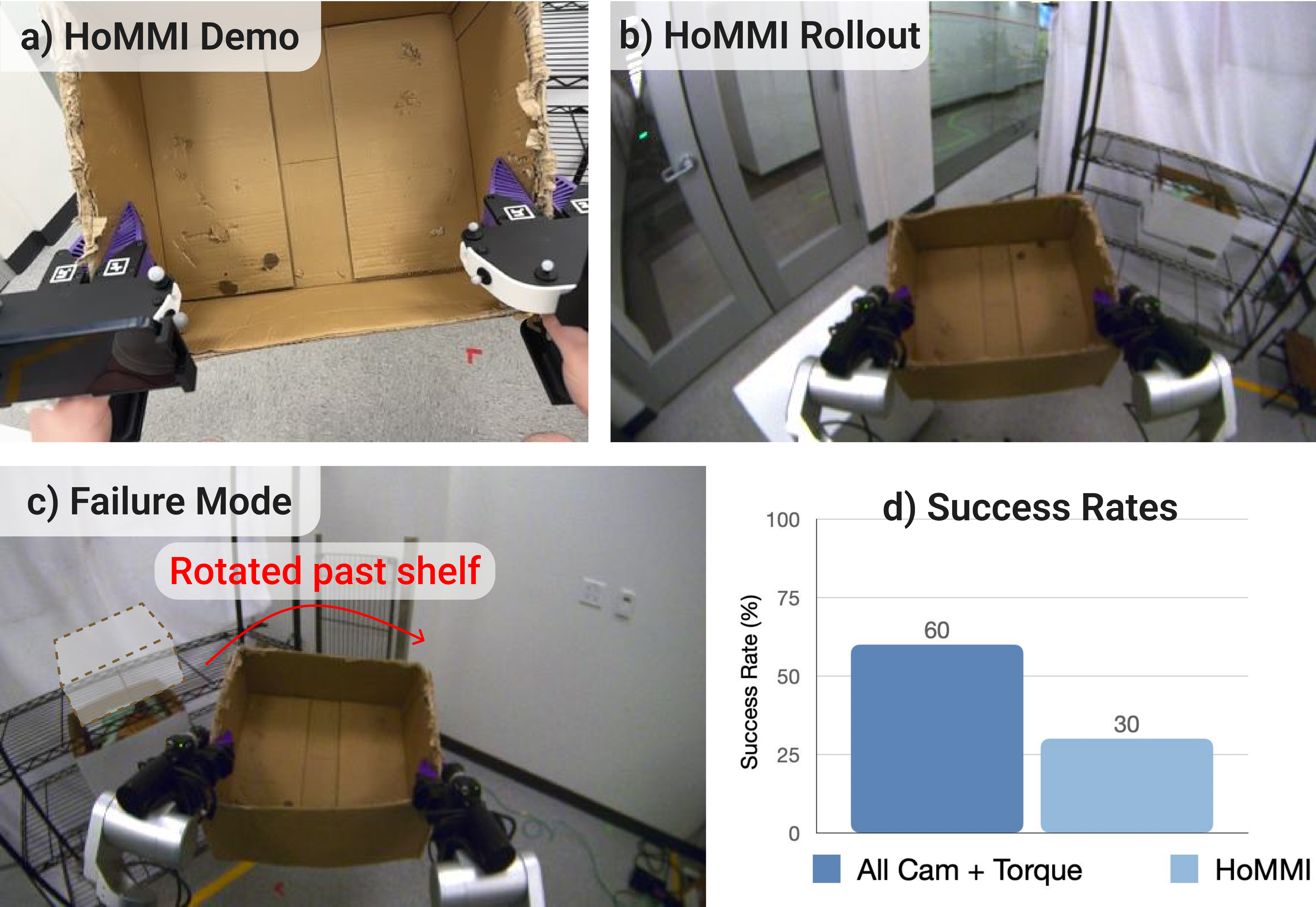}
    \caption{\footnotesize \textbf{HoMMI comparison. } a) We collect HoMMI demonstrations using two handheld grippers and the head camera visualized here. b) During rollout, there is a difference in perspective compared with demos. c) We reason that the shift in observation distribution causes the over-rotation failure. d) Compared with the All Cam + Torque ModPack policy trained with 102 demonstrations, the HoMMI policy trained on $\sim$2.17x more demonstrations achieves half the success rate.}
    \label{fig:hommi}
     \vspace{-6mm}
\end{figure}

\subsection{Comparison to alternative data collection method}
We compare against an alternative data collection approach using handheld devices, HoMMI \cite{xu2026hommilearningwholebodymobile}, which is popular due to its portability and low cost. While HoMMI enables faster data collection (40.3 s per demo with \ours~ vs. 20 s per demo with HoMMI), increased speed comes with a tradeoff in quality, as these demonstrations introduce a larger embodiment gap between data collection and deployment as shown in a) and b) in \Cref{fig:hommi}. This gap ultimately translates into lower policy performance at a similar total collection time across each dataset (68.5 minutes for \ours~and 58.9 minutes for HoMMI), suggesting that HoMMI's policy does not fully bridge the gap. 

Concretely, we trained the HoMMI policy on 221 demonstrations, approximately $2.17\times$ more than the \ours~policy. Nevertheless, the HoMMI policy succeeded in only 6/20 rollouts, compared with 12/20 for \ours. The dominant failure mode was over-rotation when the top shelf was the target placement location. We hypothesize that the larger embodiment gap makes the HoMMI policy more prone to going out of distribution at deployment. In particular, HoMMI demonstrations have a limited field of view, as shown in Figure~\ref{fig:hommi}, and the box can occlude the head-mounted camera during these examples. In contrast, \ours collects data directly on the robot, eliminating the embodiment gap between data collection and deployment.

\section{CONCLUSION}
\label{sec:conclusion}
We present \ours, a modular and extensible teleoperation system for collecting demonstrations across diverse mobile manipulation platforms and task settings. Built around a self-contained wearable backpack, \ours~provides a unified interface for plug-and-play modules including joint-level leader-arm control with haptic feedback, mobile-base control, and active perception. We demonstrate the flexibility of this design by deploying the system on two distinct robot embodiments and collecting demonstrations for real-world bimanual mobile manipulation tasks. Policies trained on data collected with \ours~achieve strong deployment performance, validating its utility as a practical data collection interface for robot learning.

\section{AI ACKNOWLEDGMENT}
For paper writing, Claude Code (e.g. Fable 5, Opus 5) was used to format \Cref{tab:leader_arms}, as well as look for missing references and reformat them in \Cref{sec:related_works}.
For codebase, Claude Code (e.g. Fable 5, Opus 5) was used to refactor ModPack software stack mentioned in \Cref{sec:software} and assist integration of policy learning and robot rollout code in \Cref{sec:policy}.

%%%%%%%%%%%%%%%%%%%%%%%%%%%%%%%%%%%%%%%%%%%%%%%%%%%%%%%%%%%%%%%%%%%%%%%%%%%%%%%%

%%%%%%%%%%%%%%%%%%%%%%%%%%%%%%%%%%%%%%%%%%%%%%%%%%%%%%%%%%%%%%%%%%%%%%%%%%%%%%%%

% \input{appendix}


\section*{ACKNOWLEDGMENT}
The authors would like to thank all REALab members as well as Chen Chen for feedback and guidance during this project. In particular, we thank Yihuai Gao and Jeff Liu for their work on robologger as well as Xiaomeng Xu, Dian Wang, and Jisang Park for their help with the RB-Y1 robot. We would also like to thank Haochen Shi for his advice on hardware and gravity compensation. This work was supported in part by the  
NSF Graduate Fellowship, NSF Award \#2143601, \#2037101, and \#2132519, Apple and Amazon. The views and conclusions contained herein are those of the authors and should not be interpreted as necessarily representing the official policies, either expressed or implied, of the sponsors.
% The preferred spelling of the word ÒacknowledgmentÓ in America is without an ÒeÓ after the ÒgÓ. Avoid the stilted expression, ÒOne of us (R. B. G.) thanks . . .Ó  Instead, try ÒR. B. G. thanksÓ. Put sponsor acknowledgments in the unnumbered footnote on the first page.

%%%%%%%%%%%%%%%%%%%%%%%%%%%%%%%%%%%%%%%%%%%%%%%%%%%%%%%%%%%%%%%%%%%%%%%%%%%%%%%%

% References are important to the reader; therefore, each citation must be complete and correct. If at all possible, references should be commonly available publications.

\bibliographystyle{IEEEtran}
\bibliography{references}

\end{document}